\documentclass[letterpaper, 10 pt, conference]{ieeeconf}  
\IEEEoverridecommandlockouts                            
\usepackage{graphics} 
\usepackage{epsfig} 
\usepackage{times} 
\usepackage{cite}
\usepackage{amsmath} 
\usepackage{amssymb}  
\usepackage{amsfonts}
\usepackage{algorithm}
\usepackage{algpseudocode}
\usepackage{graphicx}
\usepackage[hidelinks]{hyperref}
\usepackage[capitalise]{cleveref}
\usepackage{subcaption}
\usepackage{xcolor}
\usepackage{soul}
\usepackage[switch]{lineno} 
\usepackage{url}

\newtheorem{assumption}{\bf{Assumption}}

\title{\LARGE \bf
Prompt2Auto: From Motion Prompt to Automated Control via Geometry-Invariant One-Shot Gaussian Process Learning
}

\author{
Zewen Yang$^{1}$, 
Xiaobing Dai$^{2}$, 
Dongfa Zhang$^{1}$, 
Yu Li$^{1}$, 
Ziyang Meng$^{3}$, 
Bingkun Huang$^{1}$, \\
Hamid Sadeghian$^{1}$, 
Sami Haddadin$^{4}$
\thanks{$^{1}$Chair of Robotics and Systems Intelligence, Munich Institute of Robotics and Machine Intelligence, Technical University of Munich, 80333 Munich, Germany. 
$^{2}$Chair of Information-oriented Control, TUM School of Computation, Information and Technology, Technical University of Munich, 80333 Munich, Germany.
$^{3}$Dalian University of Technology, 116024 Dalian, China.
$^{4}$Mohamed bin Zayed University of Artificial Intelligence, 23201 Abu Dhabi, UAE. 
Corresponding author: Zewen Yang \textless{}{\tt\small zewen.yang@tum.de}\textgreater{}}
}

\begin{document}
\maketitle
\thispagestyle{empty}
\pagestyle{empty}

\begin{abstract}
Learning from demonstration allows robots to acquire complex skills from human demonstrations, but conventional approaches often require large datasets and fail to generalize across coordinate transformations. 
In this paper, we propose Prompt2Auto, a geometry-invariant one-shot Gaussian process (GeoGP) learning framework that enables robots to perform human-guided automated control from a single motion prompt. 
A dataset-construction strategy based on coordinate transformations is introduced that enforces invariance to translation, rotation, and scaling, while supporting multi-step predictions. 
Moreover, GeoGP is robust to variations in the user’s motion prompt and supports multi-skill autonomy.
We validate the proposed approach through numerical simulations with the designed user graphical interface and two real-world robotic experiments, which demonstrate that the proposed method is effective, generalizes across tasks, and significantly reduces the demonstration burden. 
The project page is available at \url{https://prompt2auto.github.io}
\end{abstract}

\section{Introduction}
Learning from demonstration (LfD) has become a widely adopted paradigm in robotics, offering an intuitive way for humans to transfer skills to the robots without the need for explicit programming~\cite{Argall_RAS2009_survey}. 
By observing and generalizing from human motion demonstrations, robots can acquire complex manipulation or navigation behaviors that are difficult to encode through conventional control and planning frameworks~\cite{Fang_IJIRA2019_Survey,Zare_TCYB2024_Survey}. 
This approach is particularly compelling in unstructured or dynamic environments, where manually designed policies often fail to handle the variability of the real-world conditions.

Recent progress in LfD has been dominated by deep learning methods, which excel at extracting high-dimensional representations and modeling nonlinear mappings from human demonstrations to robot policies. 
These approaches have achieved impressive results across different domains such as manipulation, grasping, and trajectory learning~\cite{Jang_CORL2022_Zero,Belkhale_CORL2023_HYDRA,Chi_RSS2023_Diffusion,Liu_ICRA2025_ForceMimic}. 
However, their data-hungry nature poses a critical limitation, which is that collecting large-scale demonstration datasets is often costly, labor-intensive, and in many cases infeasible. 
Moreover, retraining deep models to adapt to new tasks or environments introduces latency that undermines their practicality for online or adaptive robot learning~\cite{Levine_JMLR2016_End}. 

\begin{figure}[t]
    \centering
    \vspace{2mm}
    \includegraphics[width=1\linewidth]{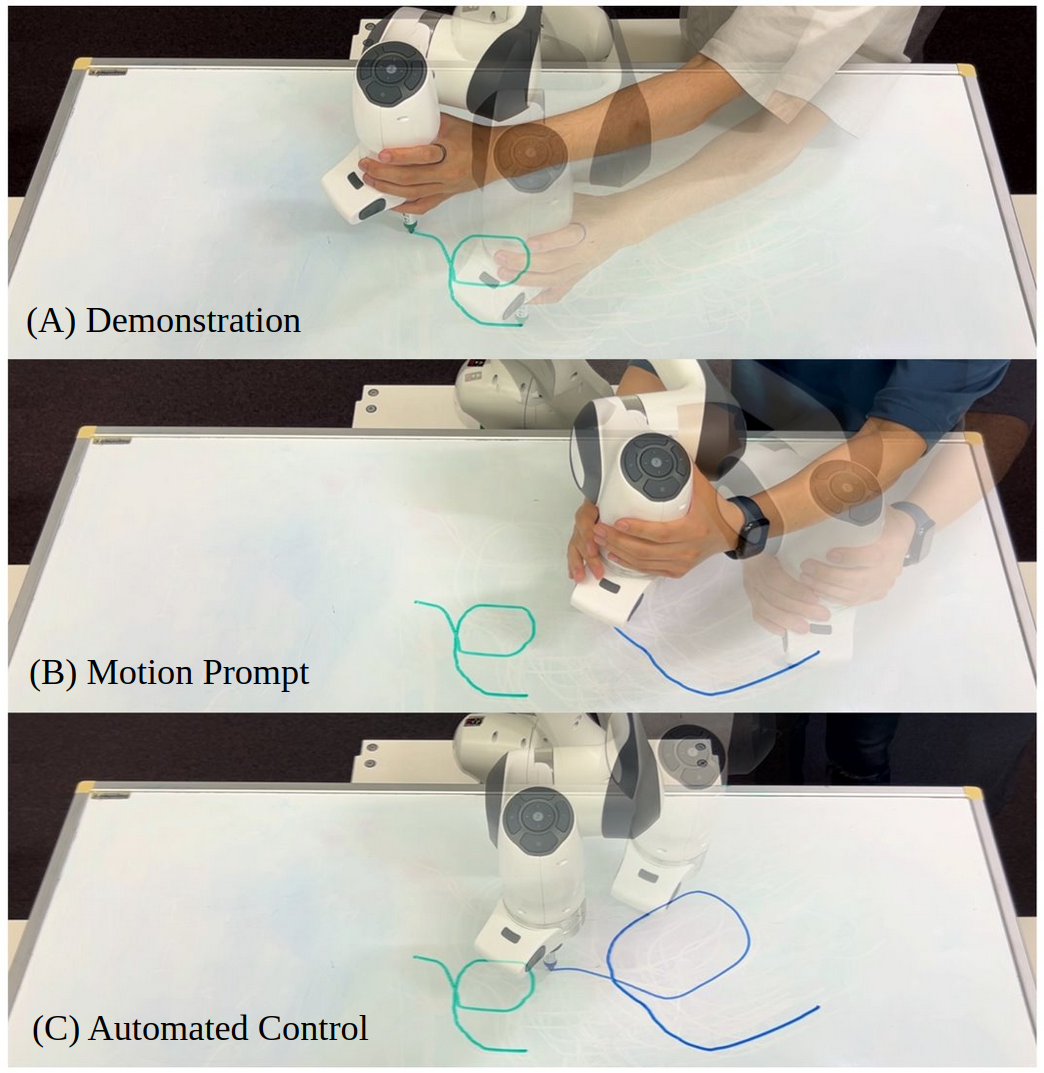}
    \caption{Illustration of the process from motion prompt to automated control: (a) the expert guides the robot along the desired reference trajectory (green), (b) the operator provides a partial motion prompt (blue), and (c) the GeoGP predicts the remaining trajectory, enabling automated robot control.}
    \label{fig_first}
\end{figure}
To address these challenges, Gaussian Process (GP) regression has emerged as a data-efficient solution for LfD. 
As a non-parametric method, it achieves strong generalization with significantly fewer samples than deep learning approaches~\cite{Rasmussen_2006_Gaussian}. 
In addition, GPs learn effectively from limited data~\cite{Dai_L4DC2023_Can}, support online adaptation through incremental updates~\cite{yang_AAAI2025_Asynchronous}, and provide calibrated uncertainty estimates that enable safe decision-making in unstructured environments~\cite{Srinivas_TIT2012_Information}.
These strengths have led to many applications in human motion prediction\cite{Wang_TPAMI2008_Gaussian,Hong_TNSRE2019_Gaussian}, pose learning with dual quaternions\cite{Lang_ICRA2014_Gaussian}, and trajectory prediction for domains such as ship navigation\cite{Rong_OE2019_Ship} and space debris tracking\cite{Yu_ASR2023_Sparse}. 
Despite these successes, existing GP-based approaches for LfD remain limited in their ability to generalize to unseen but geometrically similar motions. 
In particular, they often fail to reproduce a trajectory when its starting point is shifted or when the query trajectory exhibits variations without data argument techniques~\cite{Shorten_JBD2019_survey}, which restricts their applicability in practical scenarios. 
Furthermore, most existing GP approaches assume access to multiple demonstrations, whereas in many real-world settings, only a single demonstration is available, especially when data collection is costly.

In this paper, we introduce a novel framework for geometry-invariant one-shot GP (GeoGP) learning that enables robots to achieve human-guided automated control from a single motion prompt. 
The key idea is to leverage the Bayesian structure of GPs together with a geometry-aware formulation, ensuring that the learned motion model exhibits invariance properties critical for generalization. 
In contrast to deep learning methods, our approach requires only a single demonstration and is robust to variations in the human prompt, accounting for small deviations from the original trajectory.
This one-shot capability directly addresses the data scarcity challenge in LfD and facilitates more intuitive and scalable human–robot interaction.
The contributions of this paper are summarized as follows:
\begin{itemize}
    \item  A novel human-guided automated control framework based on GP models that learn from human demonstrations is proposed, enabling robots to autonomously identify and execute appropriate skills to complete the task with a motion prompt.
    \item The one-shot GP algorithm with coordinate transformation functions is developed that guarantees geometry-invariant multi-step predictions.
    \item A comprehensive evaluation, including simulations with a graphical user interface and two real-world robotic experiments, was conducted to demonstrate the effectiveness and robustness of the proposed approach.
\end{itemize}

\section{Preliminaries}
\subsection{Dynamic Model}
The joint dynamics of a robot manipulator with $n\in\mathbb{N}$ degrees of freedom  is modeled by
\begin{equation}\label{eq_robot_dynamics}
    \boldsymbol{M}(\boldsymbol{q})\, \ddot{\boldsymbol{q}} + \boldsymbol{C}(\boldsymbol{q}, \dot{\boldsymbol{q}})\, \dot{\boldsymbol{q}} + \boldsymbol{D}\dot{\boldsymbol{q}} + \boldsymbol{g}(\boldsymbol{q}) + \boldsymbol{d}(\boldsymbol{q}, \dot{\boldsymbol{q}}) = \boldsymbol{\tau},
\end{equation}
where the configuration is defined by the joint space $ \boldsymbol{q} \in \mathcal{C} \subset \mathbb{R}^n $, with $ \boldsymbol{M}(\boldsymbol{q}) $, $ \boldsymbol{C}(\boldsymbol{q}, \dot{\boldsymbol{q}}) $, and $ \boldsymbol{D} \in \mathbb{R}^{n \times n}$ representing the inertia, Coriolis/centrifugal, and damping matrices, respectively. The vector $ g(\boldsymbol{q}) \in \mathbb{R}^n$ captures the gravitational torques acting on the joints. 
The control torque vector is $ \boldsymbol{\tau}\in \mathbb{R}^n$.
Moreover, the vector function $\boldsymbol{d}(\boldsymbol{q}, \dot{\boldsymbol{q}}) \in \mathbb{R}^n$ encodes the external disturbances and model uncertainties.

\subsection{Task-Space Control}
\label{subsubsec_follower_control}
The manipulator is controlled in the task-space  using a Cartesian impedance controller to realize a desired translational twist with a specific scale. The corresponding joint torque command is computed as
\begin{equation}
\label{eqn:task-space-torque}
    \boldsymbol{\tau} = \boldsymbol{J}(\boldsymbol{q})^{\top}\boldsymbol{F} + \boldsymbol{N}(\boldsymbol{q})\boldsymbol{\tau}_{0},
\end{equation}
where \(\boldsymbol{J}(\boldsymbol{q}) \in \mathbb{R}^{m \times n}\) is the Jacobian matrix, $\boldsymbol{F} \in \mathbb{R}^{m}$ represents the command control  in the \(m\)-dimensional task space, and \(\boldsymbol{\tau}_{0} \in \mathbb{R}^{n}\) specifies the desired null-space torque. 
The first term of the right-hand side in \eqref{eqn:task-space-torque} projects the task-space force into joint torque, ensuring the primary task is executed, and the second term enables redundancy resolution by projecting \(\boldsymbol{\tau}_{0}\) into the null-space of the Jacobian, thereby ensuring that secondary objectives can be achieved without disturbing the primary task. 
The null-space projection matrix is defined as
\begin{align}
    \boldsymbol{N}(\boldsymbol{q}) = \boldsymbol{I} - \boldsymbol{J}(\boldsymbol{q})^{\top} (\boldsymbol{J}(\boldsymbol{q})^{\#})^{\top} \in \mathbb{R}^{n \times n},
\end{align}
where $\boldsymbol{I}$ is the identity matrix, and \(\boldsymbol{J}(\boldsymbol{q})^{\#}\) denotes the dynamically consistent generalized inverse of the Jacobian \cite{sadeghian2013task}, which is given by
\begin{align}
    \boldsymbol{J}(\boldsymbol{q})^{\#} = \boldsymbol{M}(\boldsymbol{q})^{-1} \boldsymbol{J}(\boldsymbol{q})^{\top} 
    \left( \boldsymbol{J}(\boldsymbol{q}) \boldsymbol{M}(\boldsymbol{q})^{-1} \boldsymbol{J}(\boldsymbol{q})^{\top} \right)^{-1}.
\end{align}
To achieve the desired task-space motion tracking, the impedance controller is designed as
\begin{equation}
    \boldsymbol{F} = \boldsymbol{\Lambda}(\boldsymbol{q})\boldsymbol{\ddot{x}}_{d} + \boldsymbol{\Gamma}(\boldsymbol{q},\boldsymbol{\dot{q}})\boldsymbol{\dot{x}}_{d} - \boldsymbol{K}_{d}\tilde{\boldsymbol{x}} - \boldsymbol{D}_{d}\dot{\tilde{\boldsymbol{x}}} + \boldsymbol{F}_{g}
\end{equation}
with the task-space inertia matrix, Coriolis/centrifugal matrix, and gravity vector as
\begin{align}
    \boldsymbol{\Lambda}(\boldsymbol{q}) &= \left(\boldsymbol{J(\boldsymbol{q})M(\boldsymbol{q})}^{-1}\boldsymbol{J}(\boldsymbol{q})^{\top}\right)^{-1}, \\
    \boldsymbol{\Gamma}(\boldsymbol{q},\boldsymbol{\dot q}) &= {(\boldsymbol{J}(\boldsymbol{q})^{\#})^{\top}}\boldsymbol{C}(\boldsymbol{q},\boldsymbol{\dot{q}})\boldsymbol{\dot{q}} - \boldsymbol{\Lambda}(\boldsymbol{q})\dot{\boldsymbol{J}}(\boldsymbol{q})\dot{\boldsymbol{q}}, \\
    \boldsymbol{F}_g(\boldsymbol{q}) &= {(\boldsymbol{J}(\boldsymbol{q})^{\#})^{\top}} \boldsymbol{g}(\boldsymbol{q}),
\end{align}
respectively, where $\boldsymbol{K}_{d}$ and $\boldsymbol{D}_d$ are positive definite stiffness and damping matrices. 
The task-space error $\tilde{\boldsymbol{x}} \in \mathbb{R}^{m}$ is composed of both position and orientation terms, with 
$\boldsymbol{x} = [\boldsymbol{p}^{\top}, \boldsymbol{\phi}^{\top}]^{\top}$ and corresponding desired values 
$\boldsymbol{x}_{d} = [\boldsymbol{p}_{d}^{\top}, \boldsymbol{\phi}_{d}^{\top}]^{\top}$. 
The position error is given by 
$\tilde{\boldsymbol{p}} = \boldsymbol{p} - \boldsymbol{p}_{d}$, where $\boldsymbol{p}$ denotes the current position and 
$\boldsymbol{p}_{d}$ is obtained by time integration of the commanded translational velocity 
$\boldsymbol{\nu}_{d} \in \mathbb{R}^{3}$ with respect to its initial task coordinate. 
The orientation error is encoded as
\begin{equation}
    \tilde{\boldsymbol{\phi}} = \tfrac{1}{2}\left(\boldsymbol{R}_{d}^{T}\boldsymbol{R} - \boldsymbol{R}^{T}\boldsymbol{R}_{d}\right)^{\vee},
\end{equation}
where $\boldsymbol{R}, \boldsymbol{R}_{d} \in SO(3)$ are the current and desired rotation matrices.
The operator $(\cdot)^\vee : \mathcal{SO}(3) \to \mathbb{R}^3$ extracts the vector from the skew-symmetric matrix representation in $\mathcal{SO}(3)$. 
Consequently, the closed loop dynamics under the control law with external disturbance $\boldsymbol{d}(\boldsymbol{q}, \dot{\boldsymbol{q}})$ is given by
\begin{equation}
    \boldsymbol{\Lambda}(\boldsymbol{q})\ddot{\tilde{\boldsymbol{x}}} + (\boldsymbol{\Gamma}(\boldsymbol{q},\boldsymbol{\dot{q}})+ \boldsymbol{D}_{d})\dot{\tilde{\boldsymbol{x}}} + \boldsymbol{K}_{d}\tilde{\boldsymbol{x}} = \boldsymbol{d}(\boldsymbol{q}, \dot{\boldsymbol{q}}),
\end{equation}
where the error state \(\tilde{\boldsymbol{x}}\) vanishes if \(\boldsymbol{d}(\boldsymbol{q}, \dot{\boldsymbol{q}})\) gets properly compensated. 
Furthermore, the remaining null-space behavior is designed to align with the initial posture  $\boldsymbol{q}_0$ in the null-space through 
\begin{align}
    \boldsymbol{\tau}_{0} = -\boldsymbol{K}_{N}(\boldsymbol{q} - \boldsymbol{q}_{0}) - \boldsymbol{D}_{N}\dot{\boldsymbol{q}},
\end{align}
where $\boldsymbol{K}_{N}$ and $\boldsymbol{D}_{N}$ are proper null-space stiffness and damping matrices. 
Note that the projection \(\boldsymbol{N}\) in \eqref{eqn:task-space-torque} ensures that this torque does not affect the task-space motion, as it lies in the null-space of $\boldsymbol{J}$.

\subsection{Gaussian Process Regression}
\label{subsection_GP}
Considering a general vector-valued nonlinear function $\boldsymbol{f}(\mathbf{x}): \mathcal{X} \to \mathcal{Y}$, we utilize Gaussian process regression to estimate this unknown function, where the input domain is $\mathcal{X} \subset \mathbb{R}^{d_{\mathrm{x}}}$ and the output domain is $\mathcal{Y} \subset \mathbb{R}^{d_{\mathrm{y}}}$. 
Without loss of generality, we assume the function $\boldsymbol{f}(\cdot)$ is composed of $d_{\mathrm{y}}$ independent scalar functions, i.e., $\boldsymbol{f}(\cdot) = [f_1(\cdot), \dots, f_{d_{\mathrm{y}}}(\cdot)]^\top$. 
Each component $f_i: \mathcal{X} \to \mathbb{R}$ is modeled as an independent sample from a GP. 
This assumes that each function $f_i$ lies in a Reproducing Kernel Hilbert Space (RKHS) induced by a positive definite kernel $\kappa_i(\cdot, \cdot): \mathcal{X} \times \mathcal{X} \to \mathbb{R}_{\ge0}$.
To learn the function from data, we first make the following standard assumption about the dataset.
\begin{assumption}
\label{ass_dataset}
The dataset $\mathcal{D} = \big\{ \big(\mathbf{x}^{(\iota)}, \mathbf{y}^{(\iota)} \big)\big\}_{\iota = 1, \cdots, N}$ consists of $N$ input-output pairs. The configuration of the robot $\mathbf{x}^{(\iota)}$ is measurable, and the outputs $\mathbf{y}^{(\iota)}$ are observations of the unknown function $\boldsymbol{f}$ corrupted by the noise
\begin{equation}
    \mathbf{y}^{(\iota)} = \boldsymbol{f}(\mathbf{x}^{(\iota)}) + \boldsymbol{\omega}^{(\iota)},
\end{equation}
where $\boldsymbol{\omega}^{(\iota)} \in \mathbb{R}^{d_{\mathrm{y}}}$ is the measurement noise. 
For each dimension $i \in \{1, \dots, d_{\mathrm{y}}\}$, the noise component $\omega_i^{(\iota)}$ is assumed to be bounded, i.e., $| \omega_i^{(\iota)} | \le \bar{\sigma}_{n,i}$ for a known constant $\bar{\sigma}_{n,i} \ge 0$.
\end{assumption}

For each dimension $i$, we define the dataset as $\mathcal{D}_i = \{(\mathbf{x}^{(\iota)}, \mathbf{y}_i^{(\iota)})\}_{\iota=1}^N$, where 
$\mathbf{y}_i = [\mathrm{y}_i^{(1)}, \dots, \mathrm{y}_i^{(N)}]^\top$ is the vector of training outputs for the $i$-th dimension. 
Assume the posterior distribution for $f_i$ at a new test point $\mathbf{x}^* \in \mathcal{X}$ is Gaussian~\cite{Rasmussen_2006_Gaussian}, then its mean $\mu_i(\mathbf{x}^* | \mathcal{D}_i)$ and variance $\sigma_i^2(\mathbf{x}^* | \mathcal{D}_i)$ are calculated as follows
\begin{subequations} 
\label{eqn_GP_prediction}
\begin{align}
\mu_i(\mathbf{x}^* | \mathcal{D}_i) &= \boldsymbol{k}_i(\mathbf{x}^*)^\top (\boldsymbol{K}_i + \sigma_{n,i}^2 \boldsymbol{I})^{-1} \mathbf{y}_i, \\
\sigma_i^2(\mathbf{x}^* | \mathcal{D}_i) &= \kappa_i(\mathbf{x}^*, \mathbf{x}^*) \nonumber\\
&\quad - \boldsymbol{k}_i(\mathbf{x}^*)^\top (\boldsymbol{K}_i + \sigma_{n,i}^2 \boldsymbol{I})^{-1} \boldsymbol{k}_i(\mathbf{x}^*),
\end{align}
\end{subequations}
where $\boldsymbol{k}_i(\mathbf{x}^*) = [\kappa_i(\mathbf{x}^*, \mathbf{x}^{(1)}), \dots, \kappa_i(\mathbf{x}^*, \mathbf{x}^{(N)})]^\top$ is the vector of kernel evaluations between the test point and training points, and $\boldsymbol{K}_i$ is the $N \times N$ Gram matrix with entries $[\boldsymbol{K}_i]_{js} = \kappa_i(\mathbf{x}^{(j)}, \mathbf{x}^{(s)})$.

\section{Method}

\subsection{Geometric Invariance Prediction with One-Shot GPs}
We focus on the task of multi-step trajectory prediction in a planar workspace using a one-shot GP approach, primarily for clarity of exposition.
In this setting, the GP model is trained once on a historical dataset obtained from a single demonstration provided by the user. 
The primary challenge in this one-shot setting is that standard GP methods typically require multiple demonstrations to generalize effectively and are sensitive to the absolute positions of the trajectories. 
This makes them unable to handle variations in the starting point or unseen but similar motion patterns. 
To overcome this limitation, the core idea is to move away from predicting absolute positions in a global coordinate frame. 
Instead, we represent the trajectory in polar coordinates and use GP regression to learn a mapping from historical increments (differences) to the future increments of the trajectory. 
By focusing on relative changes rather than absolute positions with coordinate transformation, the learned model becomes invariant to translation, rotation, and scaling of the trajectory, enabling robust prediction even when the motion prompt deviates from the original demonstration. 
This geometric-invariant formulation allows the system to generalize from a single human demonstration while supporting accurate multi-step prediction.

\subsubsection{Coordinate Transformation}
In order to establish the training dataset, we measure the joint configuration as $\boldsymbol{q}_{\omega} \in \mathbb{R}^n$ with
\begin{align}
    \boldsymbol{q}_{\omega} = \boldsymbol{q} + \boldsymbol{\omega}_q,
\end{align}
where $\boldsymbol{\omega}_q \in \mathbb{R}^n$ is the measurement noise, bounded by a known constant $\bar{\omega}_q \in \mathbb{R}_+$, i.e., $\| \boldsymbol{\omega}_q \| \le \bar{\omega}_q$.
The corresponding end-effector position in the task space is obtained through the forward kinematics and projection as
\begin{align}
    \mathbf{p} = \boldsymbol{\Pi}\,\mathbf{T}(\boldsymbol{q}_{\omega}),
\end{align}
where $\mathbf{T}(\cdot): \mathbb{R}^{n} \to \mathbb{R}^{3}$ denotes the forward kinematics mapping from the joint configuration to the 3D Cartesian position of the end-effector. The matrix $\boldsymbol{\Pi} \in \mathbb{R}^{2 \times 3}$ is the projection operator that maps 3D Cartesian coordinates onto the 2D task space coordinates. 
In the case of projection onto the planar surface X-Y, $\boldsymbol{\Pi}$ is given by
\begin{align}
    \boldsymbol{\Pi} = 
  \begin{bmatrix}
  1 & 0 & 0 \\
  0 & 1 & 0
  \end{bmatrix}.
\end{align}
To eliminate the translational and rotational dependency, we establish a polar coordinate system with the origin fixed at the initial observation point $\mathbf{p}(t_0)=[\mathrm{p}_{1}(t_0),\mathrm{p}_{2}(t_0)]^{\top}= \mathbf{0}_{2\times1}$ in the Cartesian coordinates, where $t_0 = 0$.
All subsequent spatial coordinates are then expressed relative to this reference frame. 
Given the discrete-time data acquisition process, the temporal index $t_k \geq 0$ is employed with $k\in\mathbb{N}_{>0}$ to denote successive observations. 
Within this polar coordinate, we define the radial distance and angular orientation as follows
\begin{align}
\label{eq_r}
r(t_k) &= \sqrt{(\mathrm{p}_1(t_k) - \mathrm{p}_1(t_0))^2 + (\mathrm{p}_2(t_k) - \mathrm{p}_2(t_0))^2}, \\
\label{eq_theta}
\theta(t_k) &= \operatorname{atan2}(\mathrm{p}_2(t_k) - \mathrm{p}_2(t_0), \mathrm{p}_1(t_k) - \mathrm{p}_1(t_0)),
\end{align}
where $r(t_k)$ represents the Euclidean distance from the initial position to the current observation at discrete time $t_k$, and $\theta(t_k)$ denotes the angular displacement measured counterclockwise from the positive $\mathrm{x}_1$-axis using the arctangent function. 
The origin polar dataset is then denoted by $\mathcal{D}_{\text{polar}} = \{(r(t_k), \theta(t_k))\}_{k\in\mathbb{N}_0}$.
The final three-dimensional feature vector is constructed through continuous angle embedding, i.e., $\boldsymbol{\xi}(t_k) = [r(t_k), \cos\theta(t_k), \sin\theta(t_k)]^\top$. 
This representation addresses the inherent periodicity of angular measurements by decomposing the angle into its sine and cosine components, thereby avoiding discontinuities at the $2\pi$ boundary while preserving the complete angular information.

To enhance the generalizability of our approach, we implement data normalization by scaling all feature values to the range. 
For the radial component, which is inherently non-negative, we employ
\begin{align}
\label{eq_tilde_r}
    \tilde{r}(t_k) = {r(t_k)}/{r}_{\text{max}}\in \tilde{\mathcal{R}} \subset \mathbb{R}_{>0},
\end{align}
where ${r}_{\text{max}}=\max_t r(t)$.
For the trigonometric components, which naturally lie in $[-1,1]$, we apply an affine transformation
\begin{align}
\label{eq_tilde_cos_sin}
    \widetilde{\cos\theta}(t_k) = \frac{\cos\theta(t_k) + 1}{2}, 
\widetilde{\sin\theta}(t_k) = \frac{\sin\theta(t_k) + 1}{2}.
\end{align}
The normalized feature vector becomes
\begin{align}
\label{eq_tilde_r_cos_sin}
    \tilde{\boldsymbol{\xi}}(t_k) = [\tilde{r}(t_k), \widetilde{\cos\theta}(t_k), \widetilde{\sin\theta}(t_k)]^\top \in \tilde{\mathcal{X}},
\end{align}
which ensures that all the normalized features are $\tilde{\xi}_i(t_k) \in [0,1]$.

\subsubsection{Dataset Construction}
Instead of utilizing absolute coordinate sequences directly, an augmented input vector is constructed that incorporates both absolute positions and related velocities. This enriches the representation of the input states for providing the Gaussian Process with more informative descriptions.
Specifically, for a given time step $t_k$, we construct the feature vector by augmenting the normalized absolute polar features with incremental components as follows
\begin{align}
\label{eq_zeta}
    \boldsymbol{\zeta}(t_k) = [\tilde{\boldsymbol{\xi}}(t_k)^\top, ~\Delta\tilde{\boldsymbol{\xi}}(t_k)^\top]^\top \in \mathbb{R}^6,
\end{align}
where $\Delta\tilde{\boldsymbol{\xi}}(t_k) :=[\Delta\tilde{r}(t_k), \Delta\widetilde{\cos\theta}(t_k), \Delta\widetilde{\sin\theta}(t_k)]^{\top}= (\tilde{\boldsymbol{\xi}}(t_k) - \tilde{\boldsymbol{\xi}}(t_{k-1}))/(t_k - t_{k-1})$ denotes the discrete-time velocity in normalized feature space and the initial value is defined as $\boldsymbol{\zeta}(t_0)= \boldsymbol{0}_{6 \times 1}$
Notably, while acceleration or higher-order derivatives could in principle be included, for practical stability and robustness, using only the position and its discrete-time increments is sufficient to achieve accurate predictions in real-world applications.

To capture temporal dependencies in the trajectory data, we employ a sliding window approach with window length $w \in \mathbb{N}$. 
The input vector $\boldsymbol{\zeta}(t_k) \in \mathbb{R}^{3w}$ is constructed by concatenating the most recent $w$ feature vectors as
\begin{align}
\label{eq_tilde_zeta}
\tilde{\boldsymbol{\zeta}}(t_{k}, w) = [ {\boldsymbol{\zeta}}(t_{k-w})^\top,  {\boldsymbol{\zeta}}(t_{k-w+1})^\top, \ldots,  {\boldsymbol{\zeta}}(t_{k-1})^\top]^\top, \!
\end{align}
where $k\ge w$. 
This concatenation results in a $3w$-dimensional input vector that encodes both the spatial characteristics and the temporal evolution of the trajectory over the preceding $w$ time steps.
Therefore, the augmented dataset is constructed as $\mathcal{D}_{\text{aug}}(t_k, w) = \{(\tilde{\boldsymbol{\zeta}}(t_i,w), \Delta\tilde{\boldsymbol{\xi}}(t_i)^\top)\}_{i \in [0,k)}$. 

To address the multi-output prediction task, we employ three independent GP models for each dimension of the target vector $\Delta\tilde{\boldsymbol{\xi}}(t_k)$.
This architecture decomposes the prediction problem into three separate regression tasks, each conditioned on the same input state vector $\tilde{\boldsymbol{\zeta}}(t_k, w)$. Specifically, $\mathcal{GP}_r$ predicts $\Delta \hat{r}(t_{k+1})$, $\mathcal{GP}_{\cos}$ predicts $\Delta \widehat{\cos\theta}(t_{k+1})$, $\mathcal{GP}_{\sin}$ predicts $\Delta \widehat{\sin\theta}(t_{k+1})$. 
Then we define $\boldsymbol{\mu}(\cdot) = [{\mu}_{\text{r}}(\cdot),{\mu}_{\cos}(\cdot),{\mu}_{\sin}(\cdot)]^{\top}$ for reconstructing the one-step absolute prediction of $\hat{\boldsymbol{\xi}}$, which are calculated as follows
\begin{align}
\label{eq_hat_xi_update}
    \hat{\boldsymbol{\xi}}(t_{k+1}) =  \tilde{\boldsymbol{\xi}}(t_{k}) + \boldsymbol{\mu}(\tilde{\boldsymbol{\zeta}}(t_k,w)|\mathcal{D}_{\text{aug}}(t_f, w)),
\end{align}
where $\tilde{\boldsymbol{\xi}}(t_{k})$ denotes the measurement at the $k$-th time step, and $t_f$ indicates the time of the last measurement included in the augmented dataset $\mathcal{D}_{\text{aug}}$.
After performing the coordinate transformation and constructing the dataset, predictions can be generated within the normalized domain $\tilde{\mathcal{X}}$. However, this approach inherently assumes that the prompt motion is on the same scale as the demonstration, which is generally infeasible because the prompt motion cannot perfectly replicate the demonstrated trajectory. 
Moreover, the user may terminate the input at any arbitrary point, meaning that the prompt motion does not necessarily cover the entire trajectory observed in the demonstration. 
Therefore, a scaling strategy is required to adapt the prediction process to varying input lengths and scales. 
To determine the scale factor, we propose a deterministic approach based on anchor points. 
Considering the angular invariance during the scaling, we define an angular sequence with $C\in\mathbb{N}$ predefined checkpoints by $\check{\Theta}= \{ \check{\theta}_i \in \mathcal{D}_{\text{polar}} | i=1,\dots, C\}$ and its corresponding radial sequence by $\check{\mathcal{R}}= \{ \check{r}_i \in \tilde{\mathcal{R}} | i=1,\dots, C \}$. 
At time step $t_k$, define the index of the last reached checkpoint as
\begin{align}
C(t_k) = \max \{ i \in \{1, \dots, C\} \mid t_{\check{\theta}_i} \le t_k \},
\end{align}
where $t_{\check{\theta}_i}$ is the time at which checkpoint $\check{\theta}_i$ occurs.  
The scale factor $\lambda$ is then computed as the average ratio of the observed to the reference radial distances up to this checkpoint, as follows
\begin{align}
\label{eq_lambda}
\lambda(t_k) = \frac{1}{C(t_k)} \sum_{i=1}^{C(t_k)} \frac{r(t_i)}{\check{r}_i}.
\end{align}
Building upon this scale factor and considering the normalization functions \eqref{eq_tilde_r} and \eqref{eq_tilde_cos_sin}, we are able to make consistent predictions within the transformed domain ${\mathcal{X}}$, which are calculated as 
\begin{align}
\label{eq_bar_r}
    \bar{r}(t_k) &= \lambda(t_k) \hat{r}(t_k), \\
\label{eq_bar_theta}
    \bar{\theta} (t_k) &= \operatorname{atan2}(2{\widehat{\sin\theta}}(t_{k}) - 1, 2{\widehat{\cos\theta}}(t_{k}) - 1).
\end{align}
Consequently, the predictions of the planar position $\hat{\mathbf{p}}(t_k)$ at time $t_k$ in Cartesian coordinates are given as,
\begin{align}
\label{eq_p_update}
    \hat{\mathbf{p}}(t_{k+1}) = \hat{\mathbf{p}}(t_k) + \bar{r}(t_k) [\cos{\bar{\theta}}, \sin{\bar{\theta}}]^{\top}.
\end{align}

\begin{algorithm}[t]
\caption{One Step Prediction with GeoGP}
\label{alg_oneshotGP}
\begin{algorithmic}[1]
\Require Window length $w$, data quantity $N$; $\mathcal{D}_{\text{aug}} = \emptyset$
\Statex \textbf{Phase 1: Data Preprocessing}
\Statex Set origin at initial position: $\mathbf{p}(t_0) = [\mathrm{p}_1(t_0), \mathrm{p}_2(t_0)]^\top$

\For{$k = 1$ to $N$}
\State    $r(t_k), \theta(t_k)$ $\gets$ \cref{eq_r,eq_theta}
 \State   $\tilde{\boldsymbol{\xi}}(t_k)$ $\gets$ \cref{eq_tilde_r,eq_tilde_cos_sin}, ${\boldsymbol{\zeta}}(t_k)$ $\gets$ \cref{eq_zeta}
\EndFor

\For{$k = w$ to $N-1$}
    \State $\mathcal{D}_{\text{aug}} = \mathcal{D}_{\text{aug}} \cup \{(\tilde{\boldsymbol{\zeta}}(t_k), \Delta\boldsymbol{\zeta}(t_{k+1}))\}$ $\gets$ \cref{eq_tilde_zeta}
\EndFor

\Statex \textbf{Phase 2: GP Model Training and Prediction}
\State $\mathcal{GP}_r, \mathcal{GP}_{\cos}, \mathcal{GP}_{\sin}$ $\gets$ Train GP on $\mathcal{D}_{\text{aug}}$

\State Scale factor after $s$-th motion prompt: $\lambda(t_s)$ $\gets$ \cref{eq_lambda}

\State  GP predictions: $\hat{\boldsymbol{\xi}}(t_{s+1})$ $\gets$ \cref{eqn_GP_prediction} and \eqref{eq_hat_xi_update} 

\Statex Convert back to Cartesian coordinates:
\State  $\bar{r}(t_{s+1}), \bar{\theta}(t_{s+1})$ $\gets$ \cref{eq_bar_r} and \eqref{eq_bar_theta}
\State $\mathbf{p}(t_{s+1})$ $\gets$ \cref{eq_p_update}

\end{algorithmic}
\end{algorithm}

\subsection{Multi-Step Prediction}
\label{subsec_error_analysis}
Given the defined dataset $\mathcal{D}_{\text{aug}}$, we are able to predict the next $l\in\mathbb{N}$ step of the motions with predicted values. 
Suppose the human-guided motion prompt ends at the $k$-th step, the next $h$-step predictions are computed recursively as
\begin{align}
\label{eq_xi_hat_h}
    \hat{\boldsymbol{\xi}}(t_{h+1}) =  \hat{\boldsymbol{\xi}}(t_{h}) + \boldsymbol{\mu}(\hat{\boldsymbol{\zeta}}(t_h, w)|\mathcal{D}_{\text{aug}}(t_f, w)),
\end{align}
for $k < h < l$, where $\hat{\boldsymbol{\xi}}(t_{h}) = [\hat{r}(t_{h}), \widehat{\cos\theta}(t_{h}), \widehat{\sin\theta}(t_{h})]^{\top}$ is obtained from \cref{eq_hat_xi_update} when $h = k+1$.
For clarity, we illustrate the construction of $\hat{\boldsymbol{\zeta}}(t_{h})$ for the case $h = k+2$ as
\begin{align}
\label{eq_zeta_hat_w}
   \! \hat{\boldsymbol{\zeta}}(t_{h}, w) = [ \tilde{\boldsymbol{\zeta}}(t_{h-w})^\top, \dots, \tilde{\boldsymbol{\zeta}}(t_{h-2})^\top,   \hat{\boldsymbol{\zeta}}(t_{h-1})^\top]^\top, \!
\end{align}
where
\begin{align}
\label{eq_zeta_hat2}
\hat{\boldsymbol{\zeta}}(t_{h-1}) = \bigl[\hat{\boldsymbol{\xi}}(t_{h-1})^\top, ~\Delta\hat{\boldsymbol{\xi}}(t_{h-1})^\top\bigr]^\top.
\end{align}
Furthermore, predictions for $h+3$ till $l$ can be obtained analogously by recursively substituting the predicted states into \cref{eq_zeta_hat2}.
While our proposed model is capable of generating an arbitrary number of future predictions, it is crucial to determine an appropriate stopping point to prevent over-extrapolation and performance degradation.
To this end, we define an event-triggered stopping criterion as
\begin{align}
l = \min \{h > k \mid E_{\sigma}(t_h) \geq \delta_{\sigma} \wedge  E_{d}(t_h) \leq \delta_{d}\},
\end{align}
where $\delta_{\sigma} = \sqrt{3}\sigma_n$ is a confidence threshold derived from the nominal observation noise level $\sigma_n$, and $\delta_d \in \mathbb{R}{>0}$ is a user-defined tolerance on positional deviation.
The event-trigger functions are given by
\begin{align}
\label{eq_E_simga}
E_{\sigma}(t_h) &= \big \|\boldsymbol{\sigma}\big(\hat{\boldsymbol{\zeta}}(t_h) \mid \mathcal{D}_{\text{aug}}(t_f,w)\big)\big\|, \\
\label{eq_E_d}
E_{d}(t_h) &= \big\|{\mathbf{p}}(t_f)-\hat{\mathbf{p}}(t_{h})\big\|,
\end{align}
where $\boldsymbol{\sigma}(\cdot) = [\sigma_{\text{r}}(\cdot), \sigma_{\cos}(\cdot), \sigma_{\sin}(\cdot)]^{\top}$ denotes the vector of GP posterior variance obtained from \eqref{eqn_GP_prediction} using $\mathcal{GP}_{\text{r}}, \mathcal{GP}_{\cos}, \mathcal{GP}_{\sin}$.
In this formulation, prediction is terminated once both the model uncertainty exceeds the predefined confidence bound and the predicted position deviates beyond the final reference point of the human demonstration.

\subsection{Skill Classification for Automated Control}
The proposed framework can learn multiple skills from human demonstrations. 
Each demonstration is recorded as a separate skill, which is learned by a GP model set $\mathbf{GP}^i =\{\mathcal{GP}_{\text{r}}^i, \mathcal{GP}_{\cos}^i, \mathcal{GP}_{\sin}^i\}$ for the $i$-th skill. 
Since all demonstrations are stored, we perform skill classification by computing the similarity score between the prompt motion and each stored GP model after obtaining the scale factor \eqref{eq_lambda}. 
Specifically, the similarity score is calculated by
\begin{align}
\label{eq_e_i}
    e_i = \sum_{j\le k} \| \mathbf{p}(t_j) - {\mathbf{p}}_i(t_j) \|,  
\end{align}
where $k$ denotes the time step at which the motion prompt terminates, $\mathbf{p}(t_j)$ is the observed position at step $j$, and ${\mathbf{p}}_i(t_j)$ is the corresponding prediction from the $i$-th demonstration dataset indicating the skill.
The GP model set corresponding to the most similar skill for prediction is then selected as
\begin{align}
    i^{\ast} = \arg\min_{i} e_i,
\end{align}
where $i^{\ast}$ denotes the index of the selected skill.
The corresponding model $\mathbf{GP}^{i^{\ast}}$ is subsequently used for trajectory prediction and control execution.
Notably, the choice of $k$ should be sufficiently large to capture distinguishing features of different skills. 
If the prompt motion is too short or insufficiently informative, the classification result may become ambiguous. 
In such cases, one could either request additional input from the user or incorporate a probabilistic weighting over the most likely skills rather than relying on a single deterministic selection.

\begin{algorithm}[t]
\caption{Skill-Aware Multi-Step Prediction with GeoGP}
\label{alg_multi_step_prediction_classification}
\begin{algorithmic}[1]
\Require 
    Prompt motion $\{\mathbf{p}(t_j)\}_{j \le k}$, 
    skill library $\{\mathbf{GP}^i\}_{i=1}^{M}$,
    dataset $\mathcal{D}_{\text{aug}}(t_f,w)$, 
    thresholds $\delta_{\sigma}, \delta_d$
\Statex \textbf{Phase 1:Skill Classification}
    \For{$i = 1$ to $M$}
        \State Compute error $e_i \gets$ \cref{eq_e_i}
    \EndFor
    \State Select skill index $i^{\ast} \gets \arg\min_i e_i$
    \State Set active model set $\mathbf{GP} \gets \mathbf{GP}^{i^{\ast}}$

\Statex \textbf{Phase 2: Multi-Step Prediction}
    \State Initialize $h \gets k$
    \While{True}
        \State Construct $\hat{\boldsymbol{\zeta}}(t_{h},w)$ using \eqref{eq_zeta_hat_w} and \eqref{eq_zeta_hat2}
        \State Compute GP prediction
        $\hat{\boldsymbol{\xi}}(t_{h+1}) \gets$ \cref{eq_xi_hat_h}
        
    \State $E_{\sigma}(t_{h+1})$ and $E_{d}(t_{h+1})$ $\gets$ \cref{eq_E_simga} and \eqref{eq_E_d} 
        \If{$E_{\sigma}(t_{h+1}) \geq \delta_{\sigma}$ \textbf{and} $E_{d}(t_{h+1}) \leq \delta_{d}$}
            \State $l \gets h+1$
            \State \textbf{break}
        \EndIf
        \State $h \gets h+1$
    \EndWhile
\end{algorithmic}
\end{algorithm}

\section{Experimental Evaluations}

\begin{figure*}[htbp]
    \centering
    \begin{subfigure}[b]{0.32\linewidth}
        \centering
        \includegraphics[width=\linewidth]{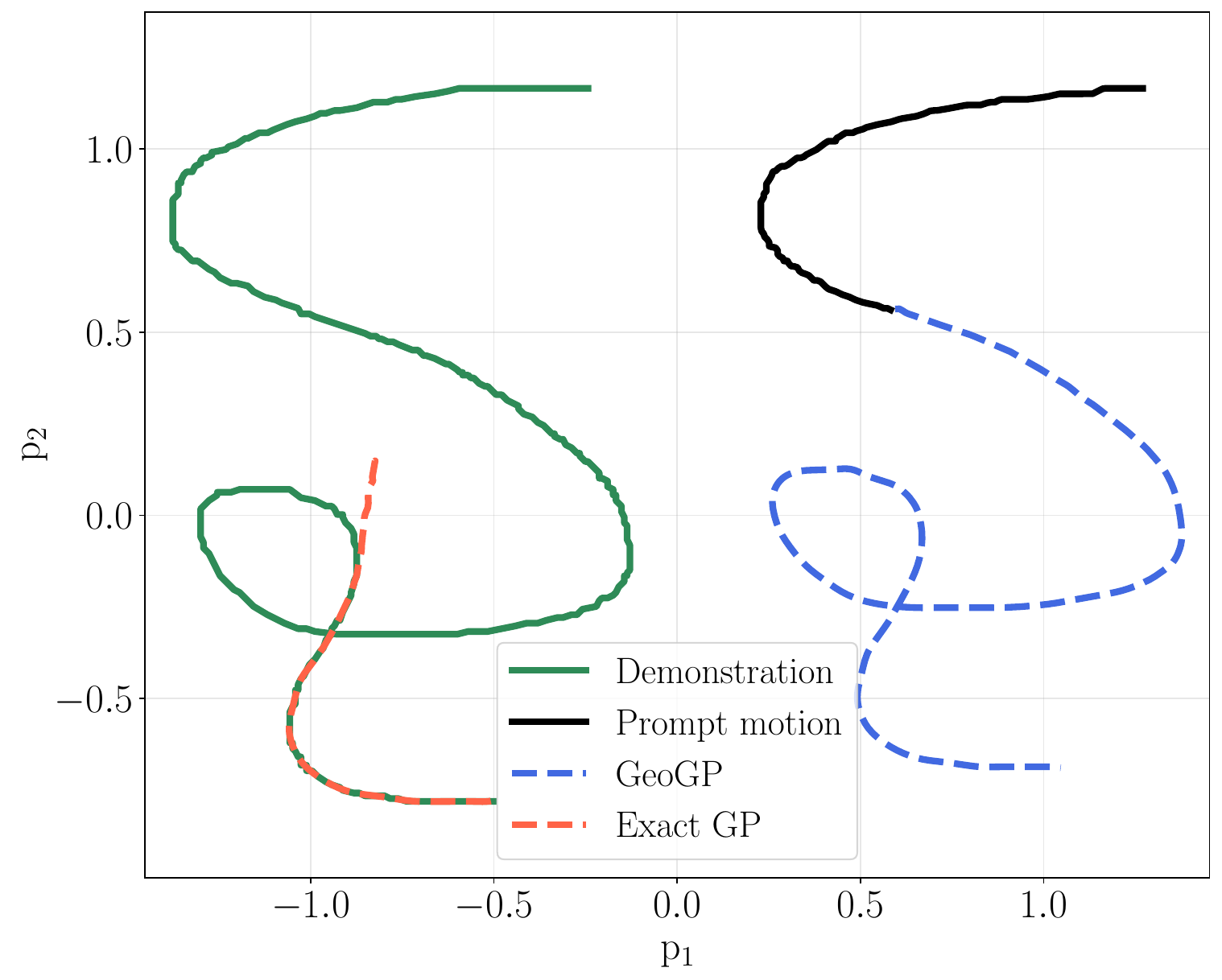}
        \caption{Translation}
        \label{fig:python_sim_translation}
    \end{subfigure}
    \hfill
    \begin{subfigure}[b]{0.32\linewidth}
        \centering
        \includegraphics[width=\linewidth]{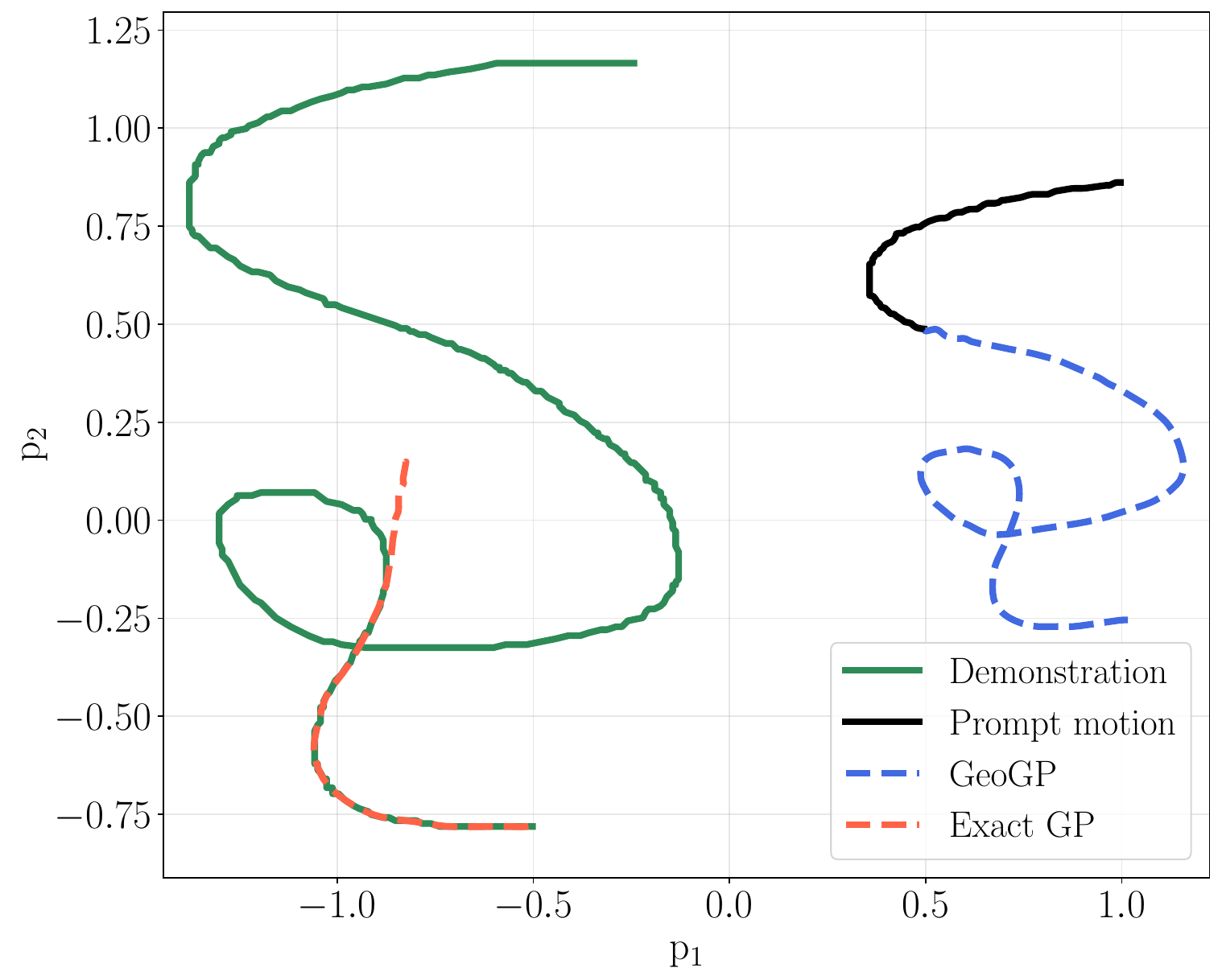}
        \caption{Scaling}
        \label{fig:python_sim_scaling}
    \end{subfigure}
    \hfill
    \begin{subfigure}[b]{0.32\linewidth}
        \centering
        \includegraphics[width=\linewidth]{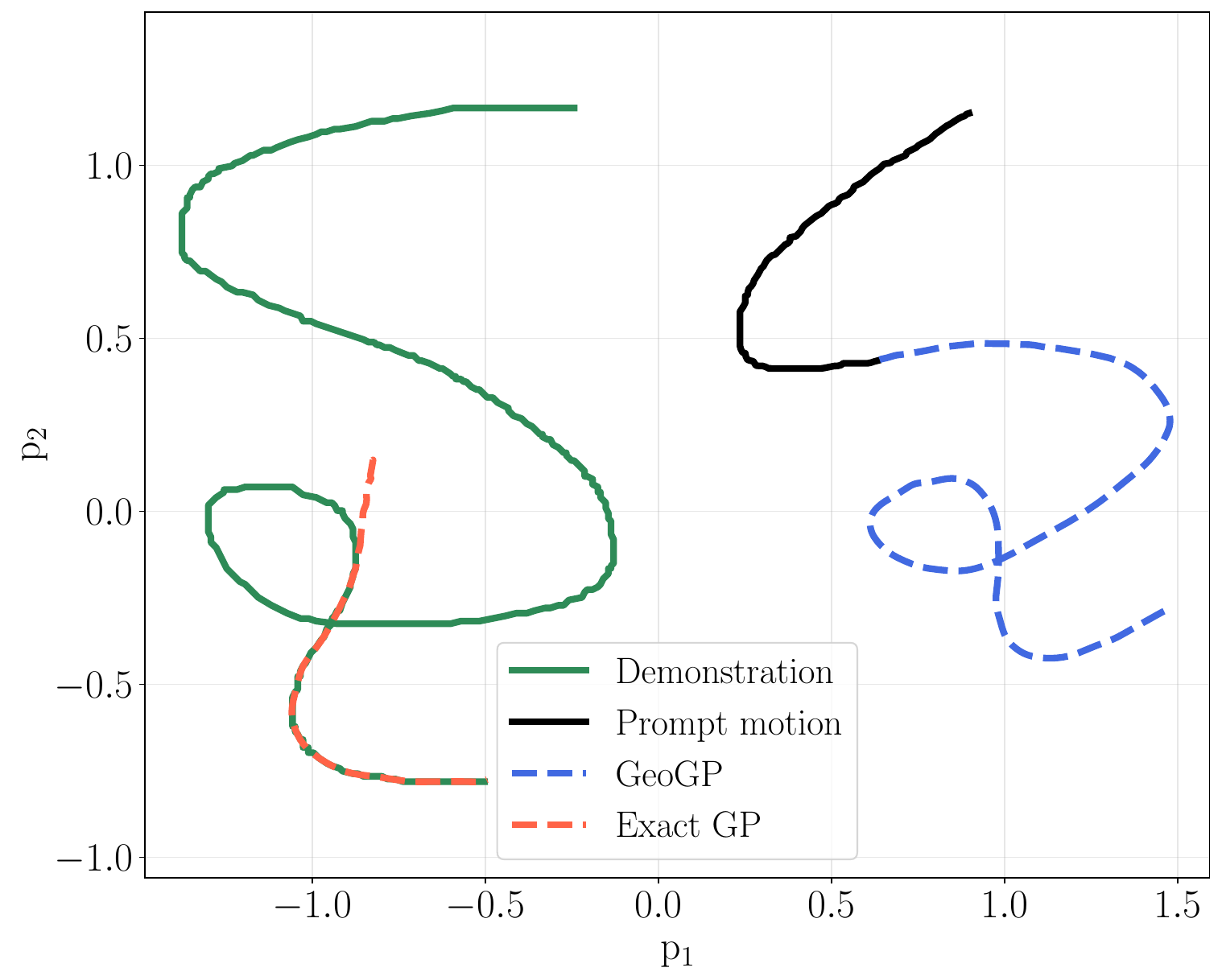}
        \caption{Rotation}
        \label{fig:python_sim_rotation}
    \end{subfigure}

    \caption{Comparison of motion prediction with different geometric transformations in simulation.}
    \label{fig:trajectory_tracking_all}
\end{figure*}
\subsection{Simulation Scenarios}
We evaluate the performance of GeoGP in a simulated environment using a graphical user interface (GUI) that allows users to draw demonstration trajectories.  
Subsequently, users provide a partially drawn trajectory as a motion prompt, simulating scenarios where only incomplete guidance is available for prediction.
The trajectory data are recorded at $20$~Hz and used to train the proposed GeoGP model. 
Following \cref{eq_tilde_zeta}, where the input vector $\boldsymbol{\zeta}(t_k) \in \mathbb{R}^{3w}$ is formed by concatenating the most recent $w$ feature vectors, we choose $w = 10$. The augmented dataset is constructed as $
\mathcal{D}_{\text{aug}}(t_k, w) = \{(\tilde{\boldsymbol{\zeta}}(t_i,w), \Delta\tilde{\boldsymbol{\xi}}(t_i)^\top)\}_{i \in [0,k)}$,
where $\tilde{\boldsymbol{\zeta}}(t_i,w) \in \mathbb{R}^{3w}$ and $\Delta\tilde{\boldsymbol{\xi}}(t_i) \in \mathbb{R}^3$. 
Accordingly, the kernel of the three independent GP models is chosen as the automatic relevance determination kernel, i.e.,
\begin{align}
k(\mathbf{x},\mathbf{x}') 
&= \sigma_f^2 \exp\!\left(
    -\tfrac{1}{2} (\mathbf{x}-\mathbf{x}')^\top 
    \boldsymbol{L}^{-1} (\mathbf{x}-\mathbf{x}')
\right),
\end{align}
where $\boldsymbol{L} = \operatorname{diag}(\ell_1^2, \ldots, \ell_{3w}^2)$, with hyperparameters set to $\ell_i=1$ ($i=1,\dots,3w$), $\sigma_f^2=1$, and $\sigma_n^2=0.01$.
After recording a reference trajectory, the user draws a partially completed probe trajectory in the GUI as a motion prompt that resembles the reference pattern but is intentionally left unfinished, as illustrated in \cref{fig:trajectory_tracking_all}. 
This mimics a scenario in which the new data are no longer available once the prompt terminates. 
At that point, the GP model performs online prediction to infer the remaining trajectory. 
This evaluation setup enables us to systematically assess the model’s ability to generalize, handle partial inputs, and adapt to geometric transformations such as translation, scaling, and rotation.  
Moreover, the predictions are compared against an exact GP model trained on Cartesian coordinates to demonstrate the advantages of the proposed geometry-invariant approach. 
Experiments are conducted under three types of geometric transformations: 
translation (\cref{fig:python_sim_translation}), 
scaling (\cref{fig:python_sim_scaling}), 
and rotation (\cref{fig:python_sim_rotation}).

To quantitatively evaluate the prediction error, we specify the reference trajectories by selecting an initial segment of the original data as motion prompts.
The prediction error is quantified as $E = \sum_{t_i \leq t_f} \| \mathbf{p}(t_i) - \hat{\mathbf{p}}(t_i) \|$, where $\mathbf{p}(t_i)$ and $\hat{\mathbf{p}}(t_i)$ denote the reference and predicted positions at time $t_i$, respectively.
\begin{figure}
    \centering
    \includegraphics[width=1\linewidth]{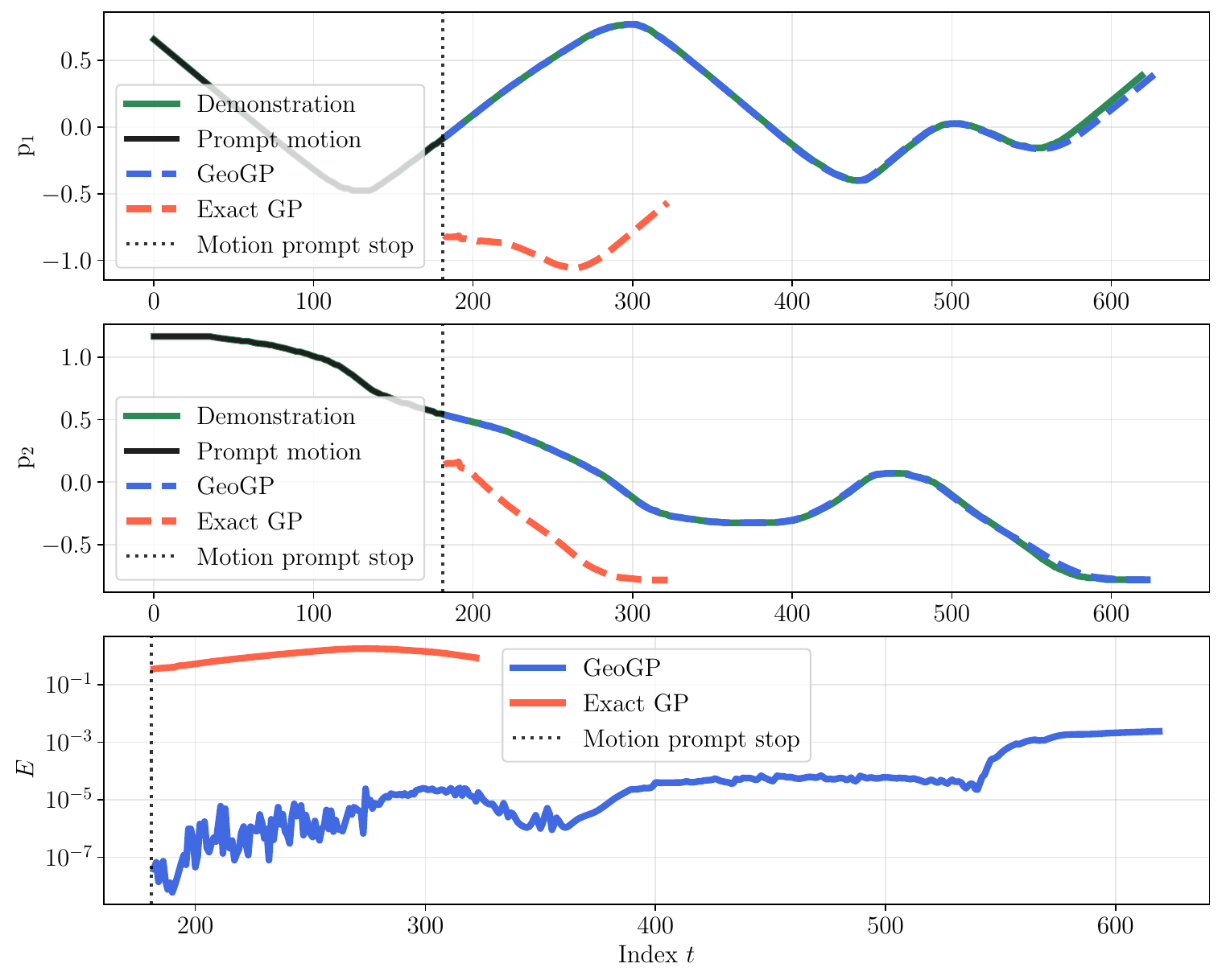}
    \caption{Motion trajectories and prediction errors.}
    \label{fig:mse}
\end{figure}
In \cref{fig:trajectory_tracking_all}, the prediction performance of GeoGP under geometric transformations against Exact GP is compared. 
The results demonstrate that GeoGP remains robust across the three considered transformations, whereas Exact GP fails to accurately predict the subsequent trajectory points from the prompt under the same conditions. 
A further comparison between GeoGP and exact GP is shown in \cref{fig:mse}, where the evaluation focuses on prediction performance along the $X$- and $Y$-axes as well as the prediction error $E$. 
GeoGP effectively follows the reference trajectory in both axes while maintaining a consistently low prediction error. 
In contrast, exact GP initially produces inaccurate predictions by collapsing toward the mean of the reference trajectory’s Cartesian coordinates, then gradually shifts toward the closest segment of the reference trajectory, and finally exhibits noticeable deviations near the end of the trajectory.

\subsection{Passive Takeover Validation - Teleoperation Case}

\begin{figure*}
    \centering
    \includegraphics[width=1\linewidth]{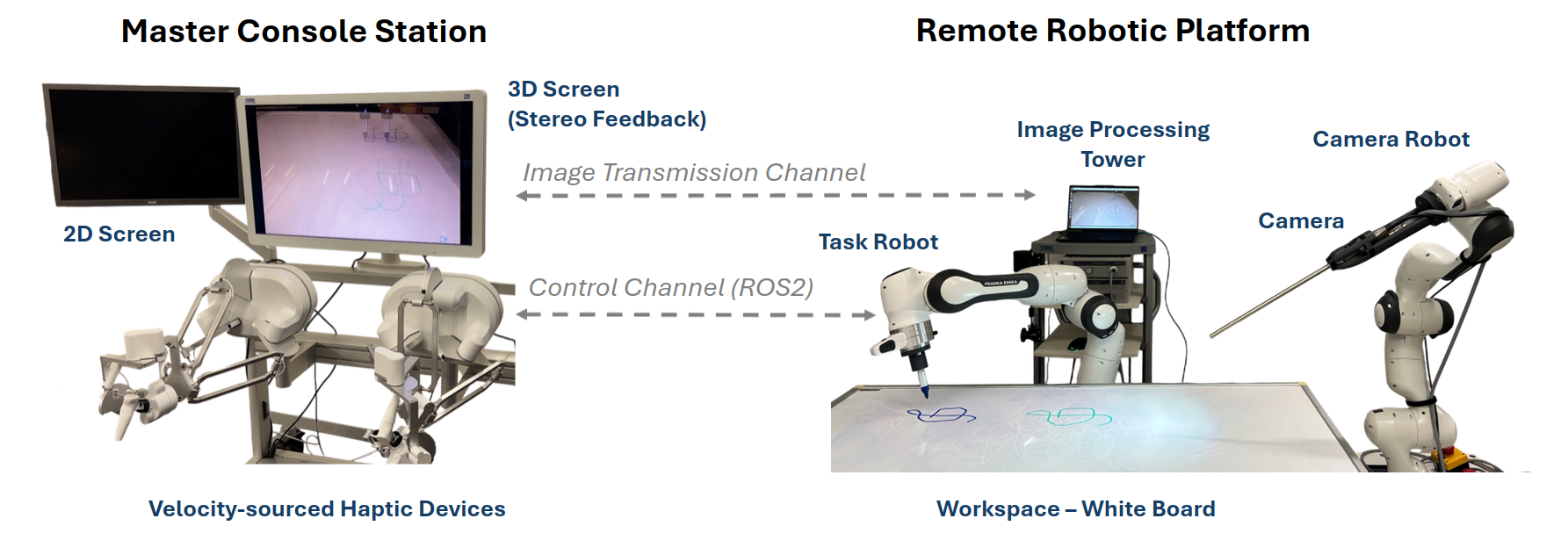}
    \caption{The overview of the teleoperated platform for the purpose of experimental evaluation.}
    \label{fig:tele-system-overview}
\end{figure*}

\begin{figure*}
    \centering
    \includegraphics[width=1\linewidth]{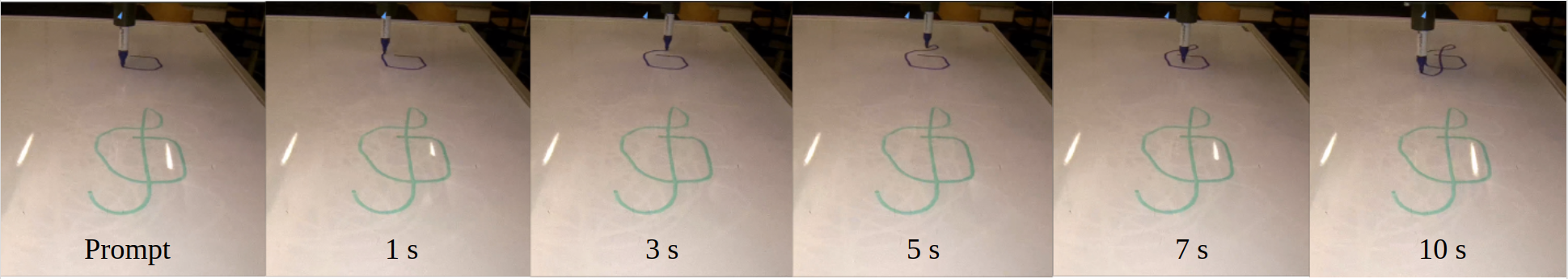}
    \caption{Validation through experiment of the symbol \textit{Treble Clef} on the teleoperated system.}
    \label{fig:exp-symbol1}
\end{figure*}

In this experiment, we validate the ability of the robot to passively take over the execution in the event of operator disengagement. 
During teleoperation, communication interruptions such as network disconnection or excessive latency can happen. 
When the remote operator is no longer able to directly teleoperate, the system automatically detects the anomaly and switches to autonomous execution of the ongoing skill. 
This scenario demonstrates the robustness of the framework, where the robot passively takes over to ensure task continuity despite teleoperation failures.

\subsubsection{Experimental Testbed}
The experimental evaluation is performed using a teleoperation testbed composed of a Master Console Station and a Remote Robotic Platform, as illustrated in \cref{fig:tele-system-overview}. The master station equips the operator with two  haptic consoles (Lambda.7 from Force Dimension) and a 3D Screen. 
The remote platform includes a 7-DoF Franka Research 3 (FR3) robot arm for task execution controlled by the left console within the workspace and another robot arm with a camera for visual acquisition, which can be controlled by the right console. 
The sensory and command data are exchanged over a communication channel built based on ROS2, bilaterally. 
The medical image processing unit from KARL STORZ, including a high-resolution Endoscope and processing tower, additionally with a video Encoder-Decoder kit from HAIVISION, is employed to capture and transfer the video signal. 
The main robot driving the task of drawing on the whiteboard is treated as a simplified and visualization-enabled scenario of a tele-surgery.

\subsubsection{Experimental Results}
We simulate a command streaming interrupt by actively interrupting the network while the operator manipulates with velocity $\boldsymbol{\nu}_{d}$ (motion of orientation is omitted for the demonstrated case) with a positive scalar gain $\gamma_{l}^{f}=10$ into the FR3’s workspace by preserving their full dynamic range and compatibility. 
As shown in~\cref{fig:exp-symbol1}, a green symbol \textit{Treble Clef} is first drawn by the task robot, controlled by the user via the Master Console Station. 
Then, GeoGP is subsequently trained from this demonstration. 
Next, a blue trajectory prompt is drawn adjacent to the demonstration. At the end of the prompt, we actively terminate the signal transmission, after which GeoGP produces a prediction and  the task robot completes the trajectory autonomously, while accounting for scaling and rotation between the demonstration and the prompt.

\subsection{Active Takeover Validation - Skill Classification Case}
\begin{figure}[h!]
    \centering
    \includegraphics[width=1\linewidth]{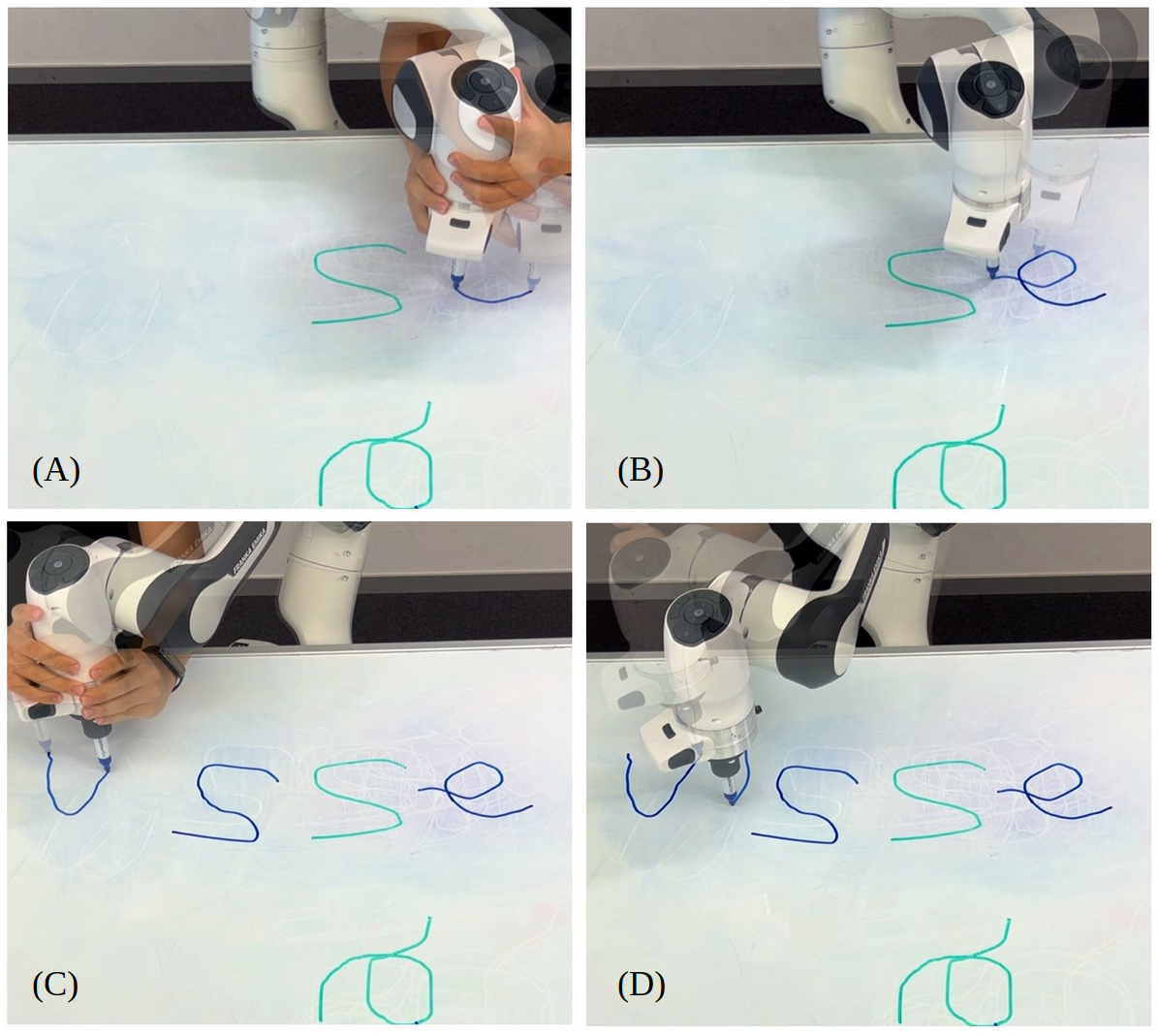}
    \caption{Illustration of the process from skill prompts to trajectory prediction. Two types of skills (denoted by the Greek letter $\alpha$ and the Latin letter $s$ in green) are first demonstrated and then prompted. (a) and (c) the expert provides partial skill segments as prompts. (b) and (d) the GeoGP model recognizes the skill types and completes the predicted motion trajectory.}
    \label{fig:enter-label}
\end{figure}
In this experiment, we validate the ability of the GeoGP to actively take over execution based on a user-provided prompt. 
The operator physically drags the robot arm along a short segment of a trajectory and then releases it. 
Using this partial demonstration, the GeoGP recognizes the corresponding skill from the learned skill library, as described in \cref{alg_multi_step_prediction_classification}. 
Once identified, the robot autonomously continues the execution from the prompted segment to complete the predicted skill motion as shown in \cref{fig:enter-label}.

\section{Discussion and Future Direction}
GPs provide data efficiency and calibrated uncertainty, yet face practical limitations for human-guided control. 
Owing to the nonparametric nature of GP models, the training and update complexity scales cubically with the number of data points, i.e., \(\mathcal{O}(N^{3})\).
But sparse or streaming approximations can mitigate this burden~\cite{Bui_NeurIPS2017_Streaming}.
When scale, rotation, and the retrieval start point are known (or reliably estimated), a compact canonicalization pipeline enables one-shot replay, e.g., anchor subtraction to form relative coordinates and rigid alignment via Procrustes superimposition. 
However, determining the start point is nontrivial, and manual specification or per-task alignment does not scale. 
To address this, GeoGP integrating invariant embeddings, phase-based representations, and few-shot adaptation will improve accuracy and robustness.

In practice, GeoGP can be embedded within a stabilizing controller, e.g., MPC, with actions gated by posterior uncertainty to ensure safe execution. 
Limitations include sensitivity to anchoring and transformation errors, and difficulties in contact-rich or discontinuous impacts even under stabilizing control. 
Future work will employ group-integrated kernels and invariant embeddings to reduce reliance on explicit canonicalization. 
Moreover, active prompting driven by GP uncertainty can be considered to elicit micro-corrections for skill classification that minimize rollout risk. 
Multi-agent skill sharing via lightweight adapters can improve scalability and reduce GP computational complexity~\cite{Yan_ACC2024_Cooperative}.
Furthermore, error compounding quantification in multi-step rollouts and explicitly enforcing geometry invariance with respect to \(\mathrm{SE}(3)\) through advanced control approaches will be investigated.

\section{Conclusion}
This paper proposed Prompt2Auto, a geometry-invariant one-shot Gaussian process learning-based framework that enables robots to perform trajectory prediction and automated control from human prompt input. 
By leveraging dataset construction based on coordinate transformation, the method achieves geometric invariance to translation, rotation, and scaling, thereby ensuring robust generalization from a single demonstration. 
Through simulations and real-world experiments, Prompt2Auto was shown to accurately predict continuous trajectories, classify skills, and seamlessly take over the tasks in both passive and active prompting scenarios. 
These results demonstrate the effectiveness of the proposed framework and showcase its potential to advance one-shot learning, improving both robustness and scalability in human-guided robot learning.


\bibliographystyle{IEEEtran}
\bibliography{ref_noURL}

\end{document}